\documentclass[11pt,a4paper]{article}
\usepackage{authblk}
\usepackage[hyperref]{emnlp2020}
\usepackage{times}
\usepackage{latexsym}

 \usepackage{array,multirow,graphicx}
 \usepackage{float}
\usepackage{footnote}

\usepackage{tikz}
\usepackage{microtype}
\usepackage{tikz-dependency}
\makesavenoteenv{tabular}
\makesavenoteenv{table}

\usepackage{amsmath}

\usepackage{verbatimbox}

\usepackage{makecell}

\usepackage[mode=buildnew]{standalone}

\aclfinalcopy 
\def\aclpaperid{826} 

\DeclareTextFontCommand{\emphb}{\bfseries\em}

\title{Persian \textit{Ezafe} Recognition Using Transformers\\ and Its Role in Part-Of-Speech Tagging}

\author[$\clubsuit$]{Ehsan Doostmohammadi}
\author[$\clubsuit$]{Minoo Nassajian}
\author[$\spadesuit$]{Adel Rahimi}

\affil[$\clubsuit$]{Sharif University of Technology, Tehran, Iran}
\affil[$\spadesuit$]{Dathena Science Pte. Ltd., Singapore}

\affil[ ]{\tt{\{e.doostm72,m.nassajian2016\}@student.sharif.edu,}}
\affil[ ]{\tt{adel.rahimi@dathena.io}}
 
\date{}

\begin{document}
\maketitle
\begin{abstract}
\textit{Ezafe} is a grammatical particle in some Iranian languages that links two words together. Regardless of the important information it conveys, it is almost always not indicated in Persian script, resulting in mistakes in reading complex sentences and errors in natural language processing tasks. In this paper, we experiment with different machine learning methods to achieve state-of-the-art results in the task of \textit{ezafe} recognition. Transformer-based methods, BERT and XLMRoBERTa, achieve the best results, the latter achieving 2.68\% F\textsubscript{1}-score more than the previous state-of-the-art. We, moreover, use \textit{ezafe} information to improve Persian part-of-speech tagging results and show that such information will not be useful to transformer-based methods and explain why that might be the case.
\end{abstract}

\section{Introduction}
\begin{figure*}[ht]
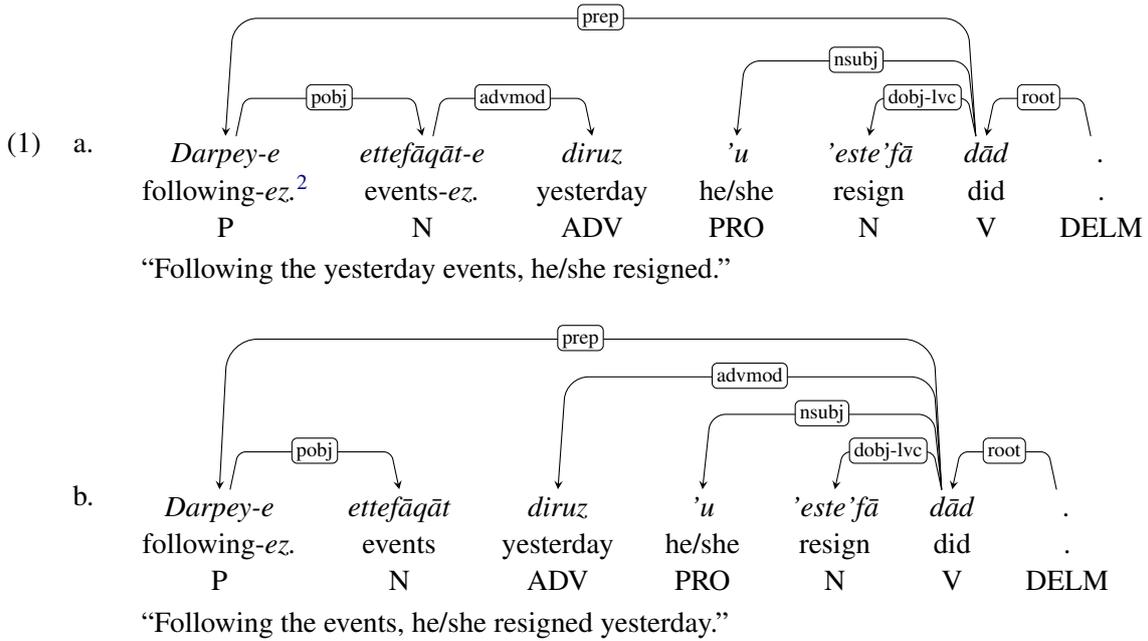

\centering

\begin{tabular}{rcl}
\addvbuffer[0ex 8ex]{(1)} & \addvbuffer[0ex 8ex]{a.} &\begin{dependency}[theme=default]
\begin{deptext}[column sep=.5cm, row sep=.1ex]
\textit{Darpey-e} \& \textit{ettef\={a}q\={a}t-e} \& \textit{diruz} \& \textit{'u} \& \textit{'este'f\={a}} \& \textit{d\={a}d} \& \textit{.} \\
following-\textit{ez.}\footnotemark \& events-\textit{ez.} \& yesterday \& he/she \& resign \& did \& . \\
P \& N \& ADV \& PRO \& N \& V \& DELM \\
\end{deptext}
\depedge{7}{6}{root}
\depedge{6}{4}{nsubj}
\depedge{6}{5}{dobj-lvc}
\depedge{1}{2}{pobj}
\depedge{2}{3}{advmod}
\depedge[edge unit distance=1.8ex]{6}{1}{prep}
\end{dependency}\\[-.5em]
&&\multicolumn{1}{l}{$\;$ ``Following the yesterday events, he/she resigned."}\\\\

& \addvbuffer[0ex 8ex]{b.} & \begin{dependency}[theme=default]
\begin{deptext}[column sep=.5cm, row sep=.1ex]
\textit{Darpey-e} \& \textit{ettef\={a}q\={a}t} \& \textit{diruz} \& \textit{'u} \& \textit{'este'f\={a}} \& \textit{d\={a}d} \& \textit{.} \\
following-\textit{ez.} \& events \& yesterday \& he/she \& resign \& did \& . \\
P \& N \& ADV \& PRO \& N \& V \& DELM \\
\end{deptext}
\depedge{7}{6}{root}
\depedge{6}{4}{nsubj}
\depedge{6}{5}{dobj-lvc}
\depedge{1}{2}{pobj}
\depedge[edge unit distance=2.3ex]{6}{1}{prep}
\depedge{6}{3}{advmod}
\end{dependency}\\[-.5em]
&&\multicolumn{1}{l}{$\;$ ``Following the events, he/she resigned yesterday."}\\
\end{tabular}

\caption{\label{fig:dep_tree}An example of the role of \textit{ezafe} in the syntactic and semantic structures.}
\end{figure*}

Persian \textit{ezafe} is an unstressed morpheme that appears on the end of the words, as \textit{-e} after consonants and as \textit{-ye}\footnote{The \textit{y} is called an intrusive \textit{y} and is an excrescence between two vowels for the ease of pronunciation.} after vowels. This syntactic phenomenon links a head noun, head pronoun, head adjective, head preposition, or head adverb to their modifiers in a constituent called `\textit{ezafe} construction' \cite{shojaei2019corpus}. Whether a word in a sentence receives or does not receive \textit{ezafe} might affect that sentence's semantic and syntactic structures, as demonstrated in Examples 1a and 1b in Figure \ref{fig:dep_tree}. There are some constructions in English that can be translated by \textit{ezafe} construction in Persian. For instance, English `of' has the same role as Persian \textit{ezafe} to show the part-whole relation, the relationship of possession, or `'s' construction, and possessive pronouns followed by nouns showing genitive cases are mirrored by Persian \textit{ezafe} \cite{karimi2012generalization}.

This affix is always pronounced but almost always not written, which results in a high degree of ambiguity in reading and understanding Persian texts. It is hence considered as one of the most interesting issues in Persian linguistics, and it has been discussed in details from phonological aspects \cite{ghomeshi1997non}, morphological aspects \cite{samvelian2006morphology, samvelian2007Ezafe} and \cite{karimi2012generalization}, and syntactic aspects
\cite{samiian1994Ezafe,larson2008Ezafe,kahnemuyipour2006persian,kahnemuyipour2014revisiting,kahnemuyipour2016Ezafe}.

Nearly 22\% of the Persian words have \textit{ezafe} \cite{bijankhan2011lessons}, which shows the prevalence of this marker. Moreover, this construction also appears in other languages such as Hawramani \cite{holmberg2005noun}, Zazaki \cite{larson2006zazaki, toosarvandani2014syntax}, Kurdish \cite{karimi2007kurdish} etc. \textit{Ezafe} construction is also similar to \textit{idafa} construction in Arabic and construct state in Hebrew \cite{habash2010introduction,karimi2012generalization} and Zulu \cite{jones2018argument}.

\textit{Ezafe} recognition is the task of automatically labeling the words ending with \textit{ezafe}, which is crucial for some tasks such as speech synthesis \cite{sheikhan1997continuous,bahaadini2011implementation}, as \textit{ezafe} is always pronounced, but rarely written. Furthermore, as recognizing the positions of this marker in sentences helps determine phrase boundaries, it highly facilitates other natural language processing (NLP) tasks, such as tokenization \cite{ghayoomi2009challenges}, syntactic parsing \cite{sagot2010morphological,nourian2015importance}, part-of-speech (POS) tagging \cite{hosseini2016persian}, and machine translation \cite{amtrup2000persian}.

In this paper, we experiment with different methods to achieve state-of-the-art results in the task of \textit{ezafe} recognition. We then use the best of these methods to improve the results for the task of POS tagging. After establishing a baseline for this task, we provide the \textit{ezafe} information to the POS tagging model once in the input text and the other time as an auxiliary task in a multi-task setting, to see the difference in the results. The contributions of this paper are (1) improving the state-of-the-art results in both of \textit{ezafe} recognition and POS tagging tasks, (2) analyzing the results of \textit{ezafe} recognition task to pave the way for further enhancement in the future work, (3) improving POS tagging results in some of the methods by providing \textit{ezafe} information and explaining why transformer-based models might not benefit from such information. The code for our experiments is available on this project's GitHub repository\footnotetext{\textit{Ezafe.}}
\footnote{\href{https://github.com/edoost/pert}{https://github.com/edoost/pert}}.

After reviewing the previous work of both tasks in Section \ref{sec:prev}, we introduce our methodology in Section \ref{sec:exp} and data in Section \ref{sec:data}. We then discuss \textit{ezafe} recognition and POS tagging tasks and their results in Sections \ref{sec:ezafe} and \ref{sec:pos}, respectively.

\section{Previous Work}
\label{sec:prev}

\subsection{\textit{Ezafe} Recognition}
\label{subsec:ezafe_literatur}
In the field of NLP, a few studies have been carried out on Persian \textit{ezafe} recognition, including rule-based methods, statistical methods, and hybrid methods. Most of the previous work on the task rely on long lists of hand-crafted rules and fail to achieve high performance on the task.

\newcite{megerdoomian2000persian} use a rule-based method to design a Persian morphological analyzer. They define an \textit{ezafe} feature to indicate the presence or absence of \textit{ezafe} for each word based on the following words in a sentence. Another work is \newcite{muller2010pergram} that considers \textit{ezafe} as a part of implemented head-driven phrase structure grammar (HPSG) to formalize Persian syntax and determine phrase boundaries. In addition, \newcite{nojoumian2011towards} designs a Persian lexical diacritizer to insert short vowels within words in sentences using finite-state transducers (FST) to disambiguate words phonologically and semantically. They use a rule-based method to insert \textit{ezafe} based on the context and the POS tags of the previous words.

As for the statistical approach, \newcite{koochari2006Ezafe} employ classification and regression trees (CART) to predict the absence or presence of \textit{ezafe} marker. They use features such as Persian morphosyntactic characteristics, the POS tags of the current word, two words before, and three words after the current word to train the model.
Their train set contains approximately 70,000 words, and the test corpus consists of 30,382 words. To evaluate the performance of the model, they use Kappa factor, and they report 98.25\% accuracy in the case of non-\textit{ezafe} words and 88.85\% in the case of words with \textit{ezafe}. As another research, we can mention \newcite{asghari2014probabilistic} that employs maximum entropy (ME) and conditional random fields (CRF) methods. They use the 10 million word Bijankhan corpus \cite{bijankhan2011lessons} and report an accuracy of 97.21\% for the ME tagger and 97.83\% for the CRF model with a window of size 5. They also utilize five Persian specific features in a hybrid setting with the models to achieve the highest accuracy of 98.04\% with CRF.

\newcite{isapour2008prediction} propose a hybrid method to determine \textit{ezafe} positions using probabilistic context-free grammar (PCFG) and then consider the relations between the heads and their modifiers. The obtained accuracy is 93.29\%, reportedly. Another work is \newcite{noferesti2014hybrid} that uses both a rule-based method and a genetic algorithm. At first, they apply 53 syntactic, morphological, and lexical rules to texts to determine words with \textit{ezafe}. Then, the genetic algorithm is employed to recognize words with \textit{ezafe}, which have not been recognized at the previous step. To train and test the model, they use the 2.5 million word Bijankhan corpus \cite{bijankhan2004persian} and obtain an accuracy of 95.26\%.

\subsection{POS Tagging}

\newcite{azimizadeh2008persian} use a trigram hidden Markov model trained on the 2.5 million word Bijankhan corpus. In order to evaluate, a variety of contexts such as humor, press reports, history, and romance are collected with 2000 words for each context. The average accuracy on different contexts is 95.11\%. 
\newcite{mohseni2010persian} also train a trigram Markov tagger on the 10 million word Bijankhan corpus. However, the lemma of each word is determined by a morphological analyzer at first and then a POS tag is assigned to the word. They report an accuracy of 90.2\% using 5-fold cross-validation on the corpus.
\newcite{hosseini2016persian} use \textit{ezafe} feature for Persian POS tagging. They use the 2.5 million word Bijankhan corpus to train a recurrent neural network-based model, whose input vectors contain the left and the right tags of the current word plus the probability of \textit{ezafe} occurrence in the adjacent words, achieving a precision of 94.7\%.
\newcite{rezai2017farsitag} design a POS tool based on a rule-based method containing both morphological and syntactic rules. They use the tag set of the 2.5 million word Bijankhan corpus, and their test set is a collection of more than 900 sentences of different types, including medicine, literature, science, etc., and the obtained accuracy is 98.6\%.
\newcite{mohtaj2018parsivar} train two POS taggers on the 2.5 million word Bijankhan corpus, ME and CRF with different window sizes, the best results of which are 95\% for both models with a window size of 5.

\section{Methodology}
\label{sec:exp}

We see both \textit{ezafe} recognition and POS tagging as sequence labeling problems, i.e., mapping each input word to the corresponding class space of the task. For the \textit{ezafe} recognition task, the class space size is two, 0 for words without and 1 for words with \textit{ezafe}. The class space size for POS tagging task is 14, consisting of the coarse-grained POS tags in the 10 million word Bijankhan corpus. The results in Section \ref{subsec:ezafe_literatur} are unfortunately reported on different, and in most cases irreproducible, test sets, using accuracy as the performance measure (which is insufficient and unsuitable for the task), making the comparison difficult. We hence re-implemented the model that reports the highest accuracy on the largest test set and compare its results with ours.

\subsection{Models}
\label{subsec:models}
We experiment with three types of models: conditional random fields (CRF) \cite{lafferty2001conditional}, recurrent neural networks (RNN) \cite{rumelhart1986learning} with long short-term memory (LSTM) cells \cite{hochreiter1997long} and convolutional neural networks (CNN), and transformer-based \cite{vaswani2017attention} models such as BERT \cite{devlin2018bert} and XLMRoBERTa \cite{conneau2019unsupervised}. These are the only transformer-based models pretrained on Persian data.
To implement these models, we used 
sklearn-crfsuite \cite{sklearncrfsuite,CRFsuite},
TensorFlow \cite{tensorflow2015-whitepaper}, PyTorch \cite{NEURIPS2019_9015},
and HuggingFace's Transformers \cite{Wolf2019HuggingFacesTS} libraries. The implementation details are as follows:

\begin{itemize}
    \item CRF\textsubscript{1}: This is a re-implementation of \newcite{asghari2014probabilistic}'s CRF model, as described in their paper. The input features were the focus word, 5 previous and 5 following words. We set the L1 and L2 regularization coefficients to $0.1$ and the max iteration argument to 100.
    
    \item CRF\textsubscript{2}: This one is the same as CRF\textsubscript{1}, plus 8 other features: 1 to 3 first and last characters of the focus word to capture the morphological information and two Boolean features indicating if the focus word is the first/last word of the sentence. 
    
    \item BLSTM: A single layer bidirectional LSTM with a hidden state size of 256 plus a fully-connected network (FCN) for mapping to the class space. The input features were Persian word embedding vectors by FastText \cite{bojanowski2017enriching} without subword information with an embedding size of 300, which is proven to yield the highest performance in Persian language \cite{8472549}. The batch size was set to 16 for \textit{ezafe} recognition and 4 for POS tagging, and learning rate to $1e-3$. We applied a dropout of rate 0.5 on RNN's output and used cross-entropy as the loss function.
    
    \item BLSTM+CNN\footnote{Number of parameters are 3.4M and 9.0M for BLSTM and BLSTM+CNN, respectively.}: The same as above, except for the input features of the BLSTM layer, which also included extracted features from dynamic character embeddings of size 32 by two CNN layers with stride 1 and kernel size 2, followed by two max-pooling layers with pool size and stride 2. The first CNN layer had 64 filters and the second one 128. We also applied a dropout of rate 0.5 on CNN's output. The character embeddings were initialized randomly and were trained with other parameters of the model.
    
    \item BERT and XLMRoBERTa: The main models plus a fully-connected network mapping to the tag space. The learning rate was set to $2e-5$ and the batch size to 8. As for the pre-trained weights, for BERT, the multilingual cased model and for XLMRoBERTa, the base model were used. We have followed the recommended settings for sequence labeling, which is to calculate loss only on the first part of each tokenized word. Cross entropy was used as the loss function.
\end{itemize}

We used Adam \cite{kingma2014adam} for optimizing all the deep models above. For \textit{ezafe} recognition, we train the models in a single-task setting. For POS tagging, however, we train them in three different settings:

\begin{enumerate}
    \item A single-task setting without \textit{ezafe} information for all of the models.
    
    \item A single-task setting with \textit{ezafe} information in the input. The outputs of the best \textit{ezafe} recognition model were added to the input of the POS tagging models: for CRFs as a Boolean feature, for BLSTM+CNN as input to CNN, and for BERT and XLMRoBERTa, in the input text. This setting was experimented with using all the models, except for CRF\textsubscript{1} and BLSTM.
    
    \item A multi-task setting where the model learns POS tagging and \textit{ezafe} recognition simultaneously, which means there is an FCN mapping to the POS class space and another one mapping to the \textit{ezafe} class space. For the BLSTM+CNN model, we used a batch size of 16 in this setting. The loss was calculated as the sum of the output losses of the two last fully-connected networks in this setting.
\end{enumerate}

The hyper-parameters of the abovementioned models have been tuned by evaluating on the validation set to get the highest F\textsubscript{1}-score. An Intel Xeon 2.30GHz CPU with 4 cores and a Tesla P100 GPU were used to train these models.

\subsection{Performance Measure}
Precision, recall, F\textsubscript{1}-score, and accuracy were used to measure the performance of each model. In all the cases, the model was tested on the test set, using the checkpoint with the best F\textsubscript{1}-score on the validation set. For the \textit{ezafe} recognition task, we report the measures on the positive class, and for the POS tagging task, we report the macro average.

\section{Data}
\label{sec:data}

The 10 million word Bijankhan \cite{bijankhan2011lessons} corpus was used in the experiments. We shuffled the corpus, as adjacent sentences might be excerpts from the same texts, with a random seed of 17 using Python's random library. This corpus comprises different topics, including news articles, literary texts, scientific textbooks, informal dialogues, etc, making it a suitable corpus for our work. We used the first 10\% of the corpus as the test, the next 10\% as validation, and the remaining 80\% as the train set. 
$\sim$22\% of the words have \textit{ezafe} marker and $\sim$78\% of them do not, in each and all of the sets. Sentences with more than 512 words were set aside. Table \ref{tab:corpus_detail} shows the number of sentences and tokens in each set.

\begin{table}[ht]
\centering
\begin{tabular}{c|cc}
\Xhline{2\arrayrulewidth}
\bf Set & \bf \# of Tokens & \bf \# of Sentences \\
\hline
Train & 8,079,657 & 268,740 \\
Valid. & 1,011,338 & 33,592 \\
Test & 1,010,274 & 33,593 \\
\hline
Total & 10,101,269 & 335,925 \\
\Xhline{2\arrayrulewidth}
\end{tabular}
\caption{\label{tab:corpus_detail} The number of sentences and tokens in train, validation, and test sets.}
\end{table}

Table \ref{tab:Ezafe_per_pos} shows the frequency percentage of \textit{ezafe} per POS in the corpus. Despite the previous claim that only nouns, adjectives, and some prepositions accept \textit{ezafe} \cite{ghomeshi1997non,karimi2012generalization,kahnemuyipour2014revisiting}, there is actually no simple categorization for POS's that accept \textit{ezafe} and those that do not, which can be seen in Table \ref{tab:Ezafe_per_pos} and is also backed by a more recent study on the matter \cite{shojaei2019corpus}.
The last column in Table \ref{tab:Ezafe_per_pos}, $H$, is Shannon's diversity index \cite{shannon1948mathematical, spellerberg2003tribute}, and is calculated as a diversity measure using Equation \ref{eq:entropy} for each POS tag. The higher the index is, the more diverse distribution the unique words have.

\begin{table}[ht]
\centering
\begin{tabular}{cccc}
\Xhline{2\arrayrulewidth}
\bf POS & \bf \% w/ \textit{Ezafe} & \bf Freq. \% & $\boldsymbol{H}$ \\
\hline
N & 46.68\% & 38.50\% & 8.518 \\
ADJ & 24.87\% & 9.02\% & 7.468 \\
P & 10.10\% & 10.90\% & 2.034 \\
DET & 9.83\% & 2.42\% & 1.944 \\
ADV & 5.67\% & 1.78\% & 5.289 \\
NUM & 2.71\% & 4.44\% & 3.573 \\
MISC & 1.59\% & 0.10\% & 3.735 \\
PRO & 1.14\% & 2.49\% & 2.884 \\
FW & 0.73\% & 0.22\% & 7.735 \\
CON & 0.12\% & 9.37\% & 1.519 \\
\hline
V & 0.00\% & 9.58\% & 5.354\\
PSTP & 0.00\% & 1.42\% & 0.029\\
IDEN & 0.00\% & 0.21\% & 3.366\\
DELM & 0.00\% & 9.54\% & 1.695\\
\Xhline{2\arrayrulewidth}
\end{tabular}
\caption{\label{tab:Ezafe_per_pos} Frequency percentage of \textit{ezafe} per POS, word frequency percentage per POS, and Shannon's diversity index ($H$) per POS.}
\end{table}

\begin{equation}
    \label{eq:entropy}
    H = -\sum_{i=1}^N P(x_i)\ln P(x_i)
\end{equation}

where $H$ is Shannon's diversity index, and $N$ is the number of unique words $x$ in each POS tag.

\section{\textit{Ezafe} Recognition}
\label{sec:ezafe}
For \textit{ezafe} recognition, we experimented with different sequence labeling techniques and report the performance of them. These techniques include CRF\textsubscript{1}, CRF\textsubscript{2}, BLSTM, BLSTM+CNN, BERT, and XLMRoBERTa, as discussed in Section \ref{subsec:models}.

\subsection{Results}
\begin{figure}[ht]
\includestandalone[width=.48\textwidth]{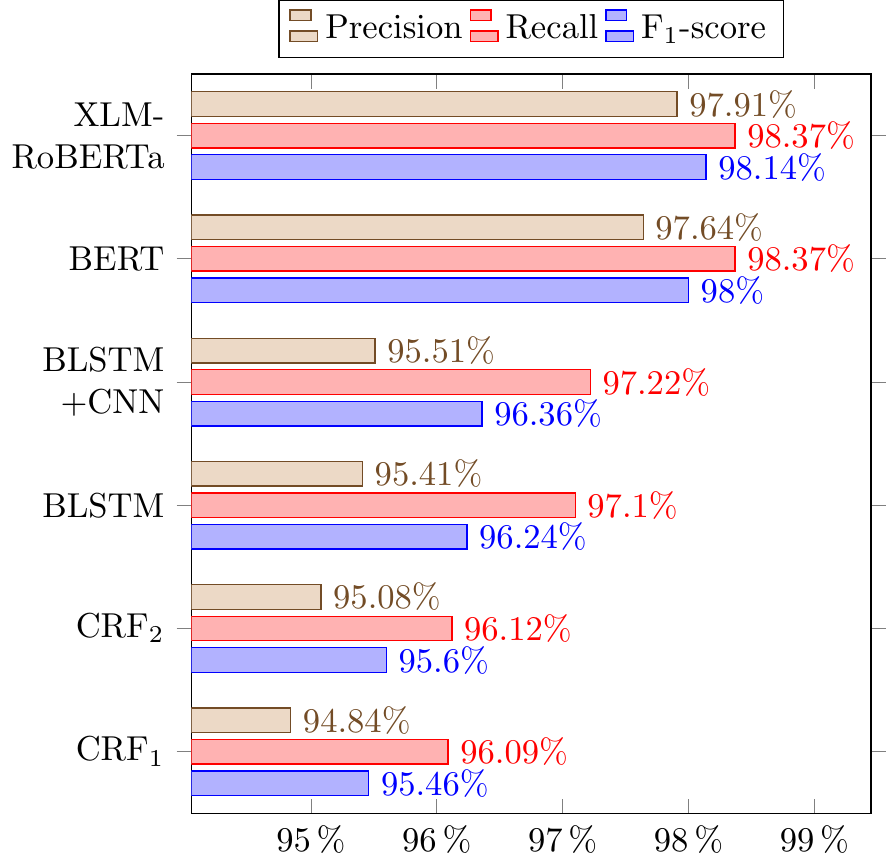}
\caption{\label{fig:ezafe_results} \textit{Ezafe} recognition precision, recall, and F\textsubscript{1}-score, respectively from top to bottom, for all of the models on the test set.}
\end{figure}

Table \ref{tab:ezafe_results} shows the results of all the models on the validation and test sets. It can be seen that transformer-based models outperform the other models by a huge margin. The best RNN-based model, BLSTM+CNN, outperforms the best CRF model, CRF\textsubscript{2}, by 0.76\% F\textsubscript{1}-score. On the other hand, the best transformer-based model, XLMRoBERTa, outperforms the best RNN by 1.78\% F\textsubscript{1}-score, and the best CRF by 2.54\%. It should be noted that XLMRoBERTa outperforms the previous state-of-the-art, CRF\textsubscript{1}, by 2.68\% F\textsubscript{1}-score. Figure \ref{fig:ezafe_results} shows the precision, recall, and F\textsubscript{1}-score on the test set. The transformer-based models also enjoy a more balanced precision and recall, which means a higher F\textsubscript{1}-score. It is worth mentioning that XLMRoBERTa has a lower training time due to its much larger pretraining Persian data compared to BERT.

\begin{table*}[t]
\centering
\small
\begin{tabular}{c|cccc|cccc|c}
\Xcline{1-10}{2\arrayrulewidth}\\[-1em]
\multirow{2}{*}{\textbf{Model}} & \multicolumn{4}{c|}{\textbf{Validation}} & \multicolumn{4}{c|}{\textbf{Test}} & \bf Approx. \\ 
& \textbf{Prec.} & \textbf{Recall} & \textbf{F\textsubscript{1}} & \textbf{Acc.} & \textbf{Prec.} & \textbf{Recall} & \textbf{F\textsubscript{1}} & \textbf{Acc.} & \bf T.T. \\ \hline
\bf CRF\textsubscript{1} (baseline) & 0.9501 & 0.9613 & 0.9556 & 0.9805 & 0.9484 & 0.9609 & 0.9546 & 0.9801 & 0.3 h \\
\bf CRF\textsubscript{2} & 0.9525 & 0.9621 & 0.9573 & 0.9812 & 0.9508 & 0.9612 & 0.9560 & 0.9807 & 0.4 h \\
\bf BLSTM & 0.9541 & 0.9712 & 0.9625 & 0.9880 & 0.9541 & \textit{0.9710} & 0.9624 & 0.9878 & 0.8 h \\
\bf BLSTM+CNN & \textit{0.9547} & \textit{0.9721} & \textit{0.9633} & \textit{0.9887} & \textit{0.9551} & \underline{0.9722} & \textit{0.9636} & \textit{0.9889} & 1 h \\
\bf BERT & \underline{0.9767} & \bf 0.9839 & \underline{0.9803} & \underline{0.9913} & \underline{0.9764} & \bf 0.9837 & \underline{0.9800} & \underline{0.9912} & 1.3 h \\
\bf XLMRoBERTa & \bf 0.9784 & \underline{0.9836} & \bf 0.9810 & \bf 0.9917 & \bf 0.9791 & \bf 0.9837 & \bf 0.9814 & \bf 0.9919 & 0.8 h \\
\Xhline{2\arrayrulewidth}
\end{tabular}
\caption{\label{tab:ezafe_results} \textit{Ezafe} recognition results (precision, recall, F\textsubscript{1}-score, and accuracy) on the validation and test sets. In each column, the best result(s) is/are in bold, the second best underlined, and the third best italicized. The last column shows the approximate training time in hours.}
\end{table*}

\subsection{Analysis}
In comparison to CRFs and RNN-based methods, transformer-based models perform much better on more scarce language forms, such as literary texts and poetry, which means, given a test corpus with a higher frequency of such texts, a much wider gap between the results is expected.
We performed an error analysis specifically on XLMRoBERTa's outputs to better understand its performance. We report \textit{ezafe} F\textsubscript{1}-score per POS tag in order of performance in Table \ref{tab:ezafe_f1_per_pos}.

\begin{table}[ht]
\centering
\begin{tabular}{cc|cc}
\Xhline{2\arrayrulewidth}
\bf POS & \bf \textit{Ezafe} F\textsubscript{1} & \bf POS & \bf \textit{Ezafe} F\textsubscript{1} \\
\hline
P & 99.78\% & NUM & 92.19\% \\
DET & 98.60\% & CON & 91.16\% \\
N & 98.14\% & PRO & 84.74\% \\
ADJ & 96.61\% & MISC & 53.85\% \\
ADV & 95.13\% & FW & 30.43\% \\
\Xhline{2\arrayrulewidth}
\end{tabular}
\caption{\label{tab:ezafe_f1_per_pos} \textit{Ezafe} F\textsubscript{1}-score per POS for XLMRoBERTa's outputs on the test set. The average F\textsubscript{1}-score is 84.06\%.}
\end{table}

\begin{itemize}
    \item Preposition (P): With a relatively low diversity and a high frequency, according to Table \ref{tab:Ezafe_per_pos}, prepositions are the easiest one to label for the \textit{ezafe} recognizing model. In addition, prepositions are exclusive in \textit{ezafe} acceptance ~93\% of the time, making this POS quite easy. The most prevalent error in this POS is the model mistaking the preposition \textit{dar} ``in" with the noun \textit{dar} ``door", the second of which accepting \textit{ezafe} almost half of the time.
    
    \item Determiners (DET): They are easy to recognize partly due to their low diversity. In this POS, the model fails to recognize \textit{ezafe} specifically when the word shares another POS in which it differs in \textit{ezafe} acceptance, e.g., \textit{hadde'aksar} ``maximum" and \textit{bi\v{s}tar} ``mostly, most of", which accept \textit{ezafe} in DET role, but not in ADV.
    
    \item Nouns (N): Despite its high diversity, the model shows high performance in detecting \textit{ezafe} in this POS. This is probably due to its high frequency and high \textit{ezafe} acceptance. Morphological information helps the most in this POS, as many nouns are derived or inflected forms of the existing words. The performance suffers from phrase boundaries detection, which results in false positives. The model also fails to recognize \textit{ezafe} on low-frequency proper nouns, such as Shakespeare. Another common error in this POS is the combination of first and last names, which are usually joined using \textit{ezafe}. 
    
    \item Adjective (ADJ) and Adverbs (ADV): Both mainly suffer from wrong detection of phrase boundaries, i.e., stopping too early or too late. For instance, look at Example 2 (the error is in bold):
    
    \begin{tabular}{llll}
        (2) & \textit{te'\={a}tr-e} & \textit{'emruz-}\emphb{e} & \textit{qarb} \\
        & theater-\textit{ez.} & contemp.-\textit{ez.} & west \\
        & \multicolumn{3}{l}{``contemporary western theater"} \\
    \end{tabular}

    \item Numbers (NUM): The errors in this POS comprise mainly the cardinal numbers, especially when written in digits. The main reason could be the scarcity of digits with \textit{ezafe}. For instance, look at Example 3 (the error is in bold):
    
    \begin{tabular}{llll}
        (3) & \textit{s\={a}l-e} & \textit{1990-}\emphb{e} & \textit{mil\={a}di} \\
        & year-\textit{ez.} & 1990-\textit{ez.} & Gregorian \\
        & \multicolumn{3}{l}{``year 1990 of the Gregorian calendar"} \\
    \end{tabular}

    \item Conjunctions (CON): It is quite rare for a conjunction to accept \textit{ezafe}, which consequently causes error in \textit{ezafe} recognition.
    
    \item Pronouns (PRO): PRO has a low \textit{ezafe} acceptance rate and a low frequency, which makes it a difficult POS. Most of the errors in this POS occur for the emphatic pronoun \textit{xod} ``itself, themselves, etc.", which receives \textit{ezafe}, as opposed to its reflective role, which does not.
    
    \item Miscellaneous (MISC): Low \textit{ezafe} acceptance and low frequency are the main reasons for the errors in this POS. The errors mainly consist of Latin single letters in scientific texts. Look at Example 4, for instance (the error is in bold):
    
    \begin{tabular}{lllll}
        (4) & \textit{L-}\emphb{e} & \textit{be} & \textit{dast} & \textit{'\={a}made} \\
        & L-\textit{ez.} & to & hand & come \\
        & \multicolumn{4}{l}{``the obtained [value of] L"} \\
    \end{tabular}

    \item Foreign words (FW): With a very low frequency, very low \textit{ezafe} acceptance rate, and a very high diversity, this POS is by far the most difficult one for the model. Additionally, FW usually appears in scientific and technical texts, which makes it harder for the model, as such texts contain a considerable amount of specialized low-frequency vocabulary. Examples of errors in this POS are `DOS', `Word', `TMA', `off', `TWTA', etc.
\end{itemize}

As discussed above, errors are most prevalently caused by model's mistaking phrase boundaries and homographs that have different syntactic roles and/or \textit{ezafe} acceptance criteria. 
While conducting the error analysis, we discovered considerable amounts of errors in Bijankhan corpus, which motivated us to correct the \textit{ezafe} labels of a part of the test corpus and measure the performance again. We, therefore, asked two annotators to re-annotate \textit{ezafe} labels of the first 500 sentences of the test corpus in parallel, and a third annotator's opinion where there is a disagreement. The results of the best model, XLMRoBERTa, on the first 500 sentences of the test corpus before and after the \textit{ezafe} label correction can be seen in Table \ref{tab:rect_results}. These 500 sentences contain 14,934 words, 3,373 of them with \textit{ezafe}, based on Bijankhan labels.

\begin{table}[ht]
\centering
\begin{tabular}{c|ccc}
\Xhline{2\arrayrulewidth}
\bf Test Corpus & \bf Precision & \bf Recall & \bf F\textsubscript{1}-score \\
\hline
\bf Bijankhan & 0.9691 & 0.9851 & 0.9770 \\
\bf Corrected & 0.9838 & 0.9897 & 0.9867 \\
\Xhline{2\arrayrulewidth}
\end{tabular}
\caption{\label{tab:rect_results} XLMRoBERTa's precision, recall, and F\textsubscript{1}-score on the first 500 sentences of the test set, before and after \textit{ezafe} label correction.}
\end{table}

Correcting \textit{ezafe} labels resulted in 0.97\% increase in F\textsubscript{1}-score on the abovementioned part of the test corpus. The same correction for all of the test corpus might result in a near 99\% F\textsubscript{1}-score for XLMRoBERTa model. Transformer-based models perform remarkably even where there is a typo crucial to \textit{ezafe} recognition, i.e., when the intrusive consonant `\textit{y}' is missed between an ending vowel and a (not-written) \textit{ezafe}, for instance, \textit{diskh\={a}-}\emphb{y}\textit{[e]} ``disks" and \textit{be'ez\={a}-}\emphb{y}\textit{[e]} ``for".

\section{POS Tagging}
\begin{table*}[th]
\centering
\small
\begin{tabular}{c|c|cccc|cccc|c}
\Xcline{1-11}{2\arrayrulewidth}\\[-1em]
&\multirow{2}{*}{\textbf{Model}} & \multicolumn{4}{c|}{\textbf{Validation}} & \multicolumn{4}{c|}{\textbf{Test}} & \bf Approx.\\ 
&& \textbf{Prec.} & \textbf{Recall} & \textbf{F\textsubscript{1}} & \textbf{Acc.} & \textbf{Prec.} & \textbf{Recall}  & \textbf{F\textsubscript{1}} & \textbf{Acc.} & \bf T.T. \\ 
\hline 
\parbox[t]{2mm}{\multirow{4}{*}{\rotatebox[origin=c]{90}{\bf Single}}}
&\bf CRF\textsubscript{1} (baseline) & 0.9688 & 0.9380 & 0.9521 & 0.9832 & 0.9680 & 0.9373 & 0.9511 & 0.9831 & 0.8 h \\
&\bf CRF\textsubscript{2} & 0.9679 & 0.9530 & 0.9602 & 0.9854 & 0.9684 & 0.9514 & 0.9595 & 0.9854 & 0.9 h \\
&\bf BLSTM+CNN & 0.9680 & 0.9573 & 0.9626 & 0.9873 & 0.9677 & 0.9570 & 0.9623 & 0.9869 & 1.3 h \\
&\bf BERT & 0.9703 & \bf 0.9719 & \underline{0.9710} & \underline{0.9899} & 0.9687 & \bf 0.9716 & \underline{0.9701} & \textit{0.9895} & 1.4 h \\
&\bf XLMRoBERTa & 0.9700 & \underline{0.9718} & \textit{0.9708} & \bf 0.9900 & 0.9706 & \underline{0.9714} & \bf 0.9709 & \bf 0.9901 & 0.9 h \\

\hline

\parbox[t]{2mm}{\multirow{4}{*}{\rotatebox[origin=c]{90}{\bf Input}}}
&\bf CRF\textsubscript{2} & 0.9697 & 0.9563 & 0.9628 & 0.9859 & 0.9708 & 0.9555 & 0.9629 & 0.9859 & 1 h \\
&\bf BLSTM+CNN & 0.9724 & 0.9597 & 0.9660 & 0.9878 & 0.9731 & 0.9587 & 0.9658 & 0.9877 & 1.4 h \\
&\bf BERT & \underline{0.9731} & \textit{0.9691} & \bf 0.9711 & \textit{0.9897} & 0.9710 & \textit{0.9690} & \textit{0.9700} & \underline{0.9897} & 1.5 h \\
&\bf XLMRoBERTa & \textit{0.9730} & 0.9689 & 0.9709 & 0.9896 & \textit{0.9714} & 0.9692 & 0.9703 & \textit{0.9895} & 1 h \\

\hline

\parbox[t]{2mm}{\multirow{3}{*}{\rotatebox[origin=c]{90}{\bf Multi}}}
&\bf BLSTM+CNN & 0.9727 & 0.9569 & 0.9647 & 0.9875 & 0.9724 & 0.9565 & 0.9643 & 0.9872 & 1.4 h \\
&\bf BERT & \bf 0.9735 & 0.9665 & 0.9699 & 0.9896 & \bf 0.9728 & 0.9650 & 0.9688 & 0.9888 & 1.5 h \\
&\bf XLMRoBERTa & \textit{0.9730} & 0.9656 & 0.9692 & 0.9887 & \underline{0.9725} & 0.9648 & 0.9686 & 0.9884 & 1 h \\\Xhline{2\arrayrulewidth}
\end{tabular}
\caption{\label{tab:pos_results} POS tagging results (precision, recall, F\textsubscript{1}-score, and accuracy) on the validation and test sets using the single- and multi-task and \textit{ezafe} in the input settings. In each column, the best result(s) is/are in bold, the second best underlined, and the third best italicized. The last column shows the approximate training time in hours.}
\end{table*}

\label{sec:pos}

For the task of POS tagging, we experimented with CRF\textsubscript{1}, CRF\textsubscript{2}, BLSTM+CNN, BERT, and XLMRoBERTa models in the single-task settings, multi-task settings with \textit{ezafe} as the auxiliary task (except for CRFs), and also in a single-task setting with \textit{ezafe} information in the input. For the last one, we added the \textit{ezafe} output of XLMRoBERTa in Section \ref{sec:ezafe} to the input text. In this section, we first explain the role of \textit{ezafe} information in POS tagging, then we discuss the results of the POS tagging task, and then we analyze it.

\subsection{The Role of \textit{Ezafe}}
\label{subsec:role_of_ezafe}
\textit{Ezafe} is a linker between words in nonverbal phrases. It is hence not used between phrases, which can be an indicator of phrase boundaries \cite{tabibzadeh2014}. Compare Examples 5a and 5b, for instance. This means that \textit{ezafe} information will help the model, and also humans, to better detect the phrase boundaries, which can be helpful in recognizing syntactic roles \cite{nourian2015importance}.

\medskip

\begin{tabular}{lllll}
    (5) & a. & [\textit{pesar}] & [\textit{xo\v{s}h\={a}l}] & [\textit{'\={a}mad}] \\
    && boy & happy & came \\
    && N & ADV & V \\
    && \multicolumn{3}{l}{``The boy came happily"} \\
    &&&&\\
    & b. & [\textit{pesar-e} & \textit{xo\v{s}h\={a}l}] & [\textit{'\={a}mad}] \\
    && boy-\textit{ez.} & happy & came \\
    && N & ADJ & V \\
    && \multicolumn{3}{l}{``The happy boy came"} \\
\end{tabular}

\medskip

Knowing \textit{ezafe} also helps the model determine the POS of some homographs. Some examples are as follows. The information below is resulted from studying homographs based on their POSs in Bijankhan corpus.

\begin{itemize}
    \item The `\textit{i}' suffix in Persian can be derivational or inflectional. When derivational, it is either a nominalizer or an adjectivizer and the derived form will accept \textit{ezafe}. When inflectional, it is an indefinite marker and the inflected form will not accept \textit{ezafe}. Some examples are \textit{kamy\={a}bi} ``scarcity, rarity", \textit{yeks\={a}ni} ``sameness", \textit{\v{s}egeft'angizi} ``wonderfulness", \textit{bim\={a}ri} ``illness", \textit{'\={a}\v{s}pazi} ``cooking".

    \item Adverbized adjectives that are homonyms in both roles, accept \textit{ezafe} only in their adjective role. For example \textit{samim\={a}ne} ``friendly, cordial" and \textit{ma'refat\v{s}en\={a}s\={a}ne} ``epistemological".
    
    \item Determiners that have a pronoun form accept \textit{ezafe} in the former, but not in the latter role. For example \textit{'aqlab} ``mostly", \textit{'aksar} ``most of", \textit{hame} ``all", \textit{'omum} ``general, most of".
    
    \item \textit{Ezafe} information might also help the model better recognize POSs that never accept \textit{ezafe}, such as verbs (V) and identifiers (IDEN).
\end{itemize}

\subsection{Results}
\begin{figure}[ht]
\includestandalone[width=.48\textwidth]{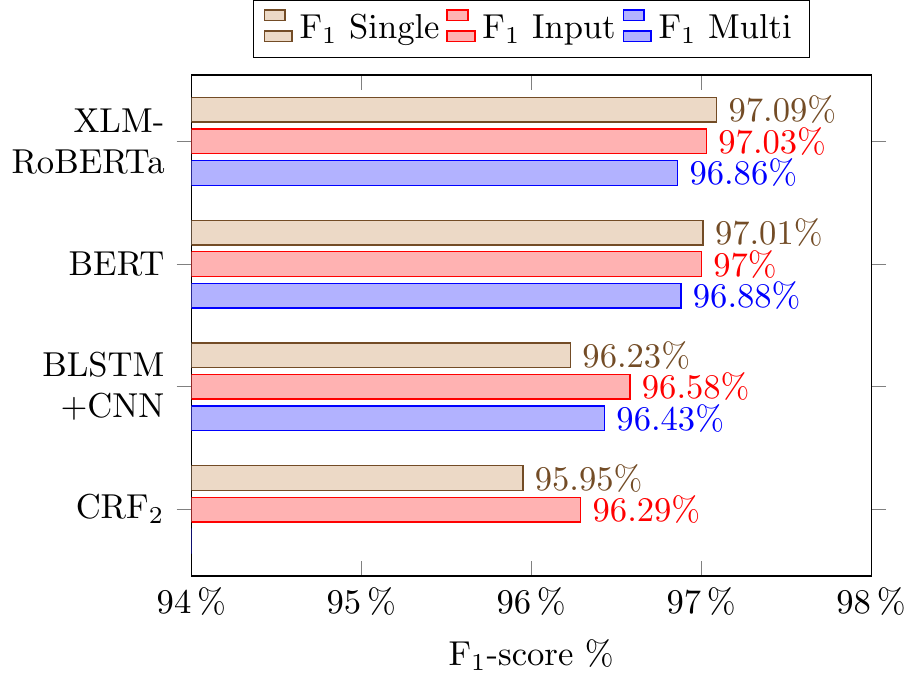}
\caption{\label{fig:pos_results} POS tagging F\textsubscript{1}-scores for single-task, input, and multi-task settings, respectively from top to bottom, on the test set.}
\end{figure}

Table \ref{tab:pos_results} shows the results of POS tagging on validation and test sets using single- and multi-task and \textit{ezafe} in the input settings. With the single-task settings, XLMRoBERTa and BERT outperform the other models and have almost equal performances. When \textit{ezafe} information is fed to the input, the precision of all the models increases while the recall shows a more complex behavior. For CRF\textsubscript{2} and BLSTM+CNN, we see a slight increase, and for the transformer-based models, we see a decrease of 0.3 to 0.4\%. The F\textsubscript{1}-score of CRF\textsubscript{2} model increases by 0.34\% and BLSTM+CNN model by 0.27\%. For BERT, it stays almost the same, and for XLMRoBERTa, it sees a decrease of 0.06\%. Table \ref{tab:pos_change} shows the change in F\textsubscript{1}-scores of each POS when \textit{ezafe} is fed with the input.
As for the multi-task settings, the precision goes up, and the recall and the F\textsubscript{1}-score come down for transformer-based and BLSTM-CNN models. Figure \ref{fig:pos_results} shows POS tagging F\textsubscript{1}-scores for single-task, in the inputs, and multi-task settings, respectively, from top to bottom, on the test set.

Table \ref{tab:pos_result_detailed} shows POS tagging F\textsubscript{1}-scores per POS on the test set for the single-task and \textit{ezafe} in the input settings for CRF\textsubscript{2} and BLSTM+CNN models and for single-task settings for XLMRoBERTa model. An increase can be seen in the F\textsubscript{1}-score when \textit{ezafe} information is provided to the model. As there is no increase in XLMRoBERTa's results when \textit{ezafe} information is provided, the results for this setting are not shown for this model.

\begin{table}[ht]
    \centering
    \small
    \addtolength{\tabcolsep}{-1pt}
    \begin{tabular}{ccc|ccc}
        \Xhline{2\arrayrulewidth}\\[-1em]
        \bf POS & \bf CRF\textsubscript{2} & \bf B.+C. & \bf POS & \bf CRF\textsubscript{2} & \bf B.+C. \\
        
        \hline
        
        \bf IDEN & 2.80\% & 2.69\% & \bf ADJ & 0.05\% & 0.07\% \\
        \bf FW & 0.79\% & 0.83\% & \bf P & 0.03\% & 0.06\% \\
        \bf ADV & 0.64\% & 0.69\% & \bf N & 0.03\% & 0.03\% \\
        \bf DET & 0.13\% & 0.16\% & \bf NUM & 0.02\% & 0.02\% \\
        \bf V & 0.06\% & 0.15\% & \bf CON & 0.01\% & 0.00\% \\
        \bf PRO & 0.06\% & 0.08\% & \bf DELM & 0.00\% & 0.00\% \\ 
        \bf MISC & 0.06\% & 0.08\% & \bf PSTP & 0.00\% & -0.01\% \\
        \Xhline{2\arrayrulewidth}
    \end{tabular}
    \addtolength{\tabcolsep}{1pt}
    
    \caption{\label{tab:pos_change} The change in POS tagging F\textsubscript{1}-scores for CRF\textsubscript{2} and BLSTM+CNN models when \textit{ezafe} information is fed with the input.}
\end{table}

\begin{table}[ht]
    \centering
    \small
    \addtolength{\tabcolsep}{-1pt}    
    \begin{tabular}{c|cccc|c}
    \Xhline{2\arrayrulewidth}\\[-1em]
        \multirow{2}{*}{\bf POS} & \multicolumn{2}{c}{\bf CRF\textsubscript{2}} & \multicolumn{2}{c|}{\bf BLSTM+CNN} & \bf X.R. \\
        & \bf Single & \bf Input & \bf Single & \bf Input & \bf Single \\
        \hline
        \bf DELM & 0.9999 & 0.9999 & 1.0000 & 1.0000 & 1.0000 \\
        \bf PSTP & 0.9995 & 0.9995 & 0.9996 & 0.9995 & 0.9998 \\
        \bf NUM & 0.9964 & 0.9966 & 0.9974 & 0.9982 & 0.9969 \\
        \bf CON & 0.9949 & 0.9950 & 0.9964 & 0.9964 & 0.9968 \\
        \bf P & 0.9944 & 0.9947 & 0.9959 & 0.9961 & 0.9966 \\
        \bf V & 0.9943 & 0.9949 & 0.9958 & 0.9964 & 0.9965 \\
        \bf N & 0.9870 & 0.9873 & 0.9893 & 0.9896 & 0.9904 \\
        \bf PRO & 0.9711 & 0.9717 & 0.9788 & 0.9795 & 0.9835 \\
        \bf DET & 0.9661 & 0.9674 & 0.9705 & 0.9713 & 0.9784 \\
        \bf ADJ & 0.9519 & 0.9524 & 0.9539 & 0.9555 & 0.9635 \\
        \bf ADV & 0.9300 & 0.9364 & 0.9414 & 0.9483 & 0.9534 \\
        \bf MISC & 0.9117 & 0.9123 & 0.9127 & 0.9142 & 0.9375 \\
        \bf FW & 0.9046 & 0.9125 & 0.9036 & 0.9119 & 0.9337 \\
        \bf IDEN & 0.8318 & 0.8598 & 0.8375 & 0.8644 & 0.8656 \\
    \Xhline{2\arrayrulewidth}
    \end{tabular}
    \addtolength{\tabcolsep}{1pt}
    
    \caption{\label{tab:pos_result_detailed} POS tagging F\textsubscript{1}-scores per POS on the test set for CRF\textsubscript{2} and BLSTM+CNN (single-task and \textit{ezafe} in the input) and for XLMRoBERTa (single-task).}
\end{table}

\subsection{Analysis}
As discussed in Subsection \ref{subsec:role_of_ezafe}, we anticipated to see an increase in several POSs, including N, ADJ, ADV, DET, V, and IDEN. According to Table \ref{tab:pos_result_detailed}, the highest increase belongs to IDEN, FW, ADV with an average increase of $\sim$2.75\%, $\sim$0.81\%, and $\sim$0.67\%, respectively. The increase for V is 0.06\% and for N, 0.03\% for both models, and for DET, 0.13\% and 0.08\%, and for ADJ, 0.05\% and 0.16\% for CRF\textsubscript{2} and BLSTM+CNN, respectively.

As for the transformer-based models results, they do not seem to benefit from the \textit{ezafe} information either in the input or as an auxiliary task. As the work on syntactic probing shows, attention heads in transformer-based models, specifically BERT, capture some dependency relation types \cite{htut2019attention}. As \textit{ezafe} is a more limited form of dependency \cite{shojaei2019corpus}, its information could be captured by the attention heads in such models. On the other hand, contextualized embeddings also seem to capture some syntactic relations \cite{47786,hewitt-manning-2019-structural}, which is another reason for such models' high performance in capturing \textit{ezafe} information. 

All in all, it seems that transformer-based models already have captured the \textit{ezafe} information owing to their architecture (attention heads), pretraining, contextual embeddings, and finally, being trained on the POS tagging task (which is related to the task of \textit{ezafe} recognition, and that is why their performance does not enhance when such information is provided.

\section{Conclusion and Future Work}
In this paper, we experimented with different models in the tasks of \textit{ezafe} recognition and POS tagging and showed that transformer-based models outperform the other models by a wide margin. We also provided \textit{ezafe} information to the POS tagging models and showed that while CRF and RNN-based models benefit from this information, transformer-based models do not. We suggest that this behavior is most probably due to (1) contextual representation, (2) pretrained weights, which means a limited knowledge of syntactic relations between words, (3) the attention heads in these models, and (4) being trained on the POS task, which is related to \textit{ezafe} recognition. An interesting direction for future work would be to investigate the role of \textit{ezafe} in transformer-based models in the tasks that such information would be helpful, such as dependency and shallow parsing.

\bibliography{emnlp2020}
\bibliographystyle{acl_natbib}

\end{document}